\documentclass[10pt,twocolumn,letterpaper]{article}

\usepackage[pagenumbers]{cvpr} % To force page numbers, e.g. for an arXiv version

\usepackage{amsmath,graphicx,amsthm,amssymb,mathrsfs,booktabs,mdwlist,multirow,colortbl,bm}
\usepackage{algorithmic,algorithm}
\newtheorem{theorem}{Theorem}
\newtheorem{lemma}{Lemma}

\begin{document}

\title{Quantile-Based Policy Optimization for Reinforcement Learning}

\author{Jinyang Jiang\\
\normalsize Department of Management Science and Information Systems, Guanghua School of Management,\\
\normalsize Peking University, Beijing, China
\and
Jiaqiao Hu\\
\normalsize Department of Applied Mathematics and Statistics,\\
\normalsize State University of New York at Stony Brook, Stony Brook, NY, USA
\and
Yijie Peng\thanks{Corresponding author: pengyijie@pku.edu.cn}\\
\normalsize Department of Management Science and Information Systems, Guanghua School of Management,\\
\normalsize Peking University, Beijing, China 
}
\maketitle

\begin{abstract}
Classical reinforcement learning (RL) aims to optimize the expected cumulative rewards. In this work, we consider the RL setting where the goal is to optimize the quantile of the cumulative rewards. We parameterize the policy controlling actions by neural networks and propose a novel policy gradient algorithm called Quantile-Based Policy Optimization (QPO) and its variant Quantile-Based Proximal Policy Optimization (QPPO) to solve deep RL problems with quantile objectives. QPO uses two coupled iterations running at different time scales for simultaneously estimating quantiles and policy parameters and is shown to converge to the global optimal policy under certain conditions. Our numerical results demonstrate that the proposed algorithms outperform the existing baseline algorithms under the quantile criterion. 
\end{abstract}

\section{Introduction}
In recent years, deep reinforcement learning (RL) has made significant achievements in games \cite{mnih2015human,silver2016mastering}, robotic control \cite{levine2016end}, recommendation \cite{zheng2018drn} and other fields. RL formulates a complex sequential decision-making task as a Markov Decision Process (MDP) and attempts to find an optimal policy by interacting with the environment. In the classical RL framework, the goal is to optimize an expected cumulative reward. However, expectation only reflects the average value of a distribution, but not the tail behavior. The tail of a distribution may capture catastrophic outcomes. For example, the joint defaults of many subprime mortgages led to the 2008 financial crisis, and in the post-crisis era, the Basel accord requires major financial institutes to maintain a minimal capital level for sustaining the loss under extreme market circumstances. In financial management, quantiles, also known as value-at-risk (VaR), can be directly translated into the minimal capital requirement. 

Risk measures have been introduced into RL either in the forms of constraints or as the objectives, which are referred as risk-sensitivity RL in the literature. The tail behavior of the distribution can be better captured by risk measures, such as VaR and conditional value-at-risk (CVaR). With risk measures as the objective function of RL, the well-trained agent can be expected to perform more robustly under extreme events.

\begin{figure}[h]
	\centering
	\includegraphics[scale=0.5]{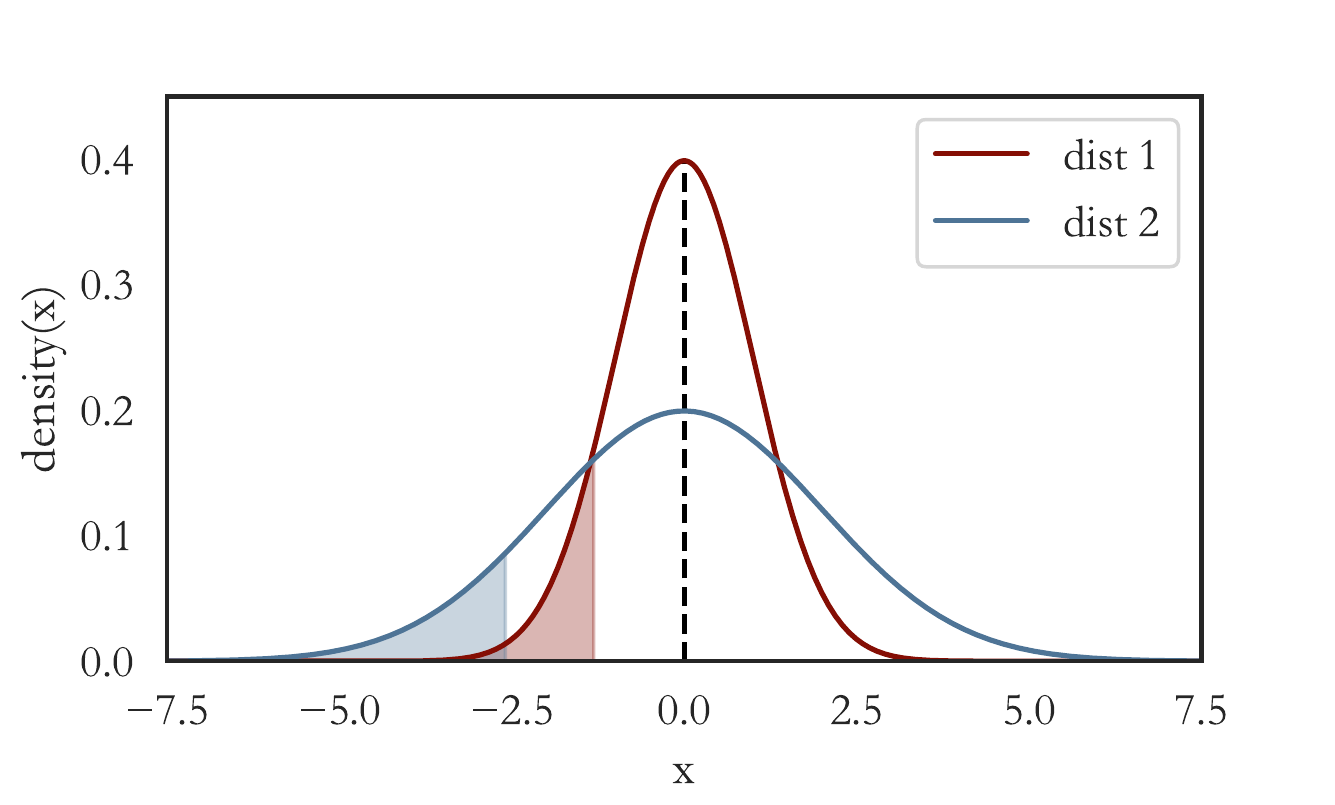}
	\caption{Probability density plots of two normal distributions $\mathcal{N}(0,1)$ and $\mathcal{N}(0,2)$. The right boundaries of shaded areas represent the value of $10\%$-quantile.}
	\label{fig:quantile_mean}
\end{figure}

In this paper, we consider an RL setting where the goal is to optimize the quantiles of the cumulative rewards. In financial investment, investors may want to obtain the highest return subject to an acceptable level of risk;  in system science, engineers may want to improve the system's most conservative output under extreme environment conditions. In Figure.\ref{fig:quantile_mean}, although the mean values of the two distributions are the same, the $10\%$-quantile of distribution 1 is significantly larger than that of distribution 2. If the returns of two portfolios follow distributions 1 and 2, then it requires more capital to avoid bankruptcy in the presence of $90\%$ possible returns for the second portfolio.
For some distributions, e.g. the Cauchy distribution, expectation may not exist so that it cannot be used as a performance measure in this situation. In contrast, quantile are always well-defined.

We parameterize the  policy controlling actions by neural networks. For RL learning problems where the dimension of the policy parameter is typically very high, gradient-based optimization provides a feasible approach. The gradient of quantiles can be expressed as a function of gradients of the distribution and quantile value. However, unlike an expectation, whose gradient can be estimated in a single simulation trajectory using techniques such as the likelihood ratio method, the gradient estimation for quantiles is much more complicated. We propose a novel policy gradient algorithm named Quantile-Based Policy Optimization (QPO) and its variant Quantile-Based Proximal Policy Optimization (QPPO) to solve deep RL problems with quantiles as objective. QPO uses two coupled iterations running at different time scales for simultaneously estimating quantiles and policy parameters and is shown to converge to the global optimal  parameterized policy under certain conditions.
Our proposed algorithms have been applied to financial investment examples, and numerical results demonstrate that the proposed algorithms outperform the existing baseline algorithms under the quantile criterion.

\section{Related Work}
\subsection{Mean-based RL}

In RL with a mean-based criterion, there are two primary categories of methods. The first category is value-based method, which learns a Q function and select the action with the best value. In deep RL, pioneer work includes Deep Q-learning (DQN) and its variants \cite{mnih2015human, hessel2018rainbow}.  The second category is based on policy optimization, where the policy is parameterized by certain basis functions and optimized by stochastic gradient ascent. REINFORCE is an early policy gradient algorithm \cite{williams1992simple} .
Trust Region Policy Optimization (TRPO) introduces the importance sampling technique into RL to improve data utilization efficiency \cite{schulman2015trust}. 
By simplifying TRPO, Proximal Policy Optimization (PPO) achieves a significant improvement through optimizing a clipped surrogate objective and has become a commonly used baseline algorithm at present \cite{schulman2017proximal}.
There are also some algorithms that combine the two categories, such as Deep Deterministic Policy Gradient (DDPG) and Soft Actor-Critic (SAC) \cite{lillicrap2015continuous, haarnoja2018soft}.

\subsection{Gradient Estimation of Risk Measure}
Gradient estimation of risk measures has been studied actively. 
Classic gradient estimation problem considers expectation \cite{fu2006gradient}, whereas 
estimating gradients of quantile and CVaR is much more complicated.  To address the difficulties, different methods, such as infinite perturbation analysis \cite{hong2009estimating,jiang2015estimating}, kernel estimation \cite{liu2009kernel, hong2009simulating}, and measure-valued differentiation \cite{heidergott2016measure}, have been developed.
Recently, \cite{glynn2021computing} uses generalized likelihood ratio estimators to estimate the gradient for a rather general distortion risk measure.
However, these methods would require analytic forms of transition probabilities or rewards in the MDP, which is typically not satisfied in the RL setting.
A well-known black-box gradient estimation approach is Simultaneous Perturbation Stochastic Approximation (SPSA) \cite{spall1992multivariate}. However, tuning parameters such as the perturbation size and step size in optimization require care and it is typically hard to apply SPSA to high-dimensional optimization such as our deep RL problems where the policy is parameterized by neural networks. 

\subsection{Risk-sensitive RL}
Risk-sensitive RL has recently gained steam. One can use a risk measure as the objective function of RL. For example, expected exponential utility is used as the objective in \cite{borkar2001sensitivity}; \cite{petrik2012approximate} and \cite{tamar2014policy} have studied CVaR-based objectives; 
\cite{prashanth2013actor} aims to optimize several
variance-related risk measures by SPSA and a smooth function approach;
\cite{prashanth2016cumulative} applies SPSA to optimize an objective function in the cumulative prospect theory.

One can also impose a risk measure as the constraint of a RL problem.
For example, dynamic and time-consistent risk constraints are considered in \cite{chow2013stochastic}; \cite{borkar2014risk} uses CVaR as the constraint; \cite{chow2017risk} develops policy gradient and actor-critic algorithms under VaR and CVaR constraints. A Lagrangian approach has been used to solve the RL problem subject to some risk measures  \cite{bertsekas1997nonlinear}. Tuning a multi-scale iterative algorithm to update Lagrange multipliers requires care. 

To the best of our knowledge, our work is the first to propose a two-time-scale iterative algorithm to optimize quantiles, which can be applied to large-scale optimization for policy  parameterized by neural networks in deep RL problems.

The rest of the paper is organized as follows: In Section 3, we describe the MDP settings, the classical framework of policy gradient algorithms, and our quantile optimization framework for RL. Section 4 introduces our new algorithms called QPO and its variant QPPO. In Section 5, we establish the  convergence to the global optimum for QPO. Numerical reuslts are presented in Section 6. We conclude the paper and discuss future direction in Section 7.

\section{Problem Formulation and Preliminaries}
\subsection{Markov Decision Process}
A Markov Decision Process (MDP) can be represented as a 5-tuple $(\mathcal{S},\mathcal{A},p,u,\eta)$, where $\mathcal{S}$ and $\mathcal{A}$ are state and action spaces, respectively; $p(s'|s,a)$ is the transition probability; $u(s',a,s)$ is the reward function; $\eta\in(0,1)$ is the reward discount factor. One may expect to optimize a parameterized policy function $\pi(a|s;\theta)$, which is a distribution of the action conditional on the state, through the interaction with the MDP system. We denote the state and action at time $t\in\{0,1,2,\cdots,T\}$ by $s_t$ and $a_t$, where $s_t\sim p(\cdot |s_{t-1}, a_{t-1})$ and $a_t\sim\pi(\cdot| s_{t};\theta)$. Then the trajectory generated by the MDP can be defined as $\tau=\{s_0,a_0,s_1,\cdots,a_{T-1},s_T\} \sim \Pi(\cdot;\theta)$. Here we have 
$\Pi(\tau;\theta)=p(s_0)\prod_{t=0}^{T-1}\pi(a_t| s_{t};\theta)p(s_{t+1}|s_{t},a_{t})$,
where $s_0$ is the initial state. The total reward of the trajectory is denoted as $R=\sum_{t=0}^{T-1}\eta^t u(s_{t+1},a_{t},s_{t})=U(\tau)$, which follows the distribution $ F_R(\cdot;\theta)$.  In an episode, a complete simulation trajectory of the action and state is generated by the interaction between the agent and environment.

\subsection{Mean-based Criterion for RL}
In the classical setting, the objective function of RL algorithms is to maximize the expected cumulative reward:
\begin{align}
    \max\limits_{\theta\in\Theta}\mathbb{E}_{R\sim F_R(\cdot;\theta)}[R]=\max\limits_{\theta\in\Theta}\mathbb{E}_{\tau\sim\pi(\cdot;\theta)}[U(\tau)].\label{pf.eq1}
\end{align}
Policy gradient methods, an important class of RL algorithms, solve problem (\ref{pf.eq1}) by the stochastic gradient ascent. The key for implementing policy gradient methods is to estimate gradient, and the likelihood ratio method is a popular stochastic gradient estimation technique. Specifically, the gradient can be expressed as below:
\begin{align}
    \nabla_{\theta}\mathbb{E}[U(\tau)]&=\nabla_{\theta}\int_{\Omega_\tau}U(\tau)\Pi(\tau;\theta)d\tau\nonumber\\
    &=\int_{\Omega_\tau}U(\tau)\frac{\nabla_{\theta}\Pi(\tau;\theta)}{\Pi(\tau;\theta)}\Pi(\tau;\theta)d\tau\nonumber\\
    &=\mathbb{E}\bigg[U(\tau)\sum_{t=0}^{T-1}\nabla_{\theta}\log\pi(a_t|s_t;\theta)\bigg], \label{pf.eq2}
\end{align}
 where the interchange of the gradient and integral in the second equality can be justified by the dominated convergence theorem. The term  $U(\tau)\sum_{t=0}^{T-1}\nabla_{\theta}\log\pi(a_t|s_t;\theta)$ inside the expectation on the right hand side of (\ref{pf.eq2}) is an unbiased stochastic gradient estimation of $\nabla_{\theta}\mathbb{E}[U(\tau)]$.

\subsection{Quantile-based Criterion for RL}
For a given probability level $\alpha\in(0,1)$, the $\alpha$-quantile for distribution $F_R(\cdot;\theta)$ is defined as
\begin{align*}
    q(\alpha;\theta)=\arg\inf\{r:P(R(\theta)\leq r)=F_R(r;\theta)\geq\alpha\}.
\end{align*}
We assume that $F_R(r;\theta)$ is continuously differentiable on $\mathbb{R}$, i.e. $F_R(r;\theta)\in C^1(\mathbb{R})$, so that the $\alpha$-quantile can be written as an inverse of the distribution function $q(\alpha;\theta)=F_R^{-1}(\alpha;\theta)$.
Our goal is to maximize the $\alpha$-quantile of the distribution on a compact convex set $\Theta\subset\mathbb{R}^m$, i.e.,
\begin{align}
    \max\limits_{\theta\in\Theta}q(\alpha;\theta)=\max\limits_{\theta\in\Theta}F^{-1}_R(\alpha;\theta). \label{pf.eq3}
\end{align}
To solve problem (\ref{pf.eq3}), we consider a stochastic gradient ascent method. By definition,  $F_R(q(\alpha;\theta);\theta)=\alpha$. Taking gradients on both sides with respect to $\theta$, we have
\begin{align}
    \nabla_{\theta} q(\alpha;\theta)=-\frac{\nabla_{\theta} F_R(r;\theta)}{f_R(r;\theta)}\bigg|_{r=q(\alpha;\theta)}. \label{pf.eq4}
\end{align}
There are two difficulties for obtaining a single-run unbiased stochastic gradient estimator:
\begin{itemize}
\item[$\bullet$] The right hand side of  equality (\ref{pf.eq4}) contains the $\alpha$-quantile itself. The quantile would change in the sequential update of $\theta$.
\end{itemize}
\begin{itemize}
\item[$\bullet$] The density function $f_R(\cdot;\theta)$ of the cumulative reward usually does not have an analytical form. We have $f_R(r;\theta)=\frac{\partial}{\partial r}F_R(r;\theta)=\frac{\partial}{\partial r}\mathbb{E}[\mathbf{1}\{R\leq r\}]$, but the gradient and expectation operators cannot be interchanged since $\mathbf{1}\{R\leq r\}$ is discontinuous.
\end{itemize}

\section{Quantile-Based Policy Optimization}
In this section, we propose a new on-policy RL algorithm, named Quantile-based Policy Optimization (QPO) algorithm and its variant Quantile-based Proximal Policy Optimization (QPPO) algorithm to solve the MDP with quantile criterion (\ref{pf.eq3}). The computation complexity of our algorithms will also be discussed. 

\subsection{Quantile Optimization of REINFORCE Style}

Since the denominator of the $\alpha$-quantile in equality (\ref{pf.eq4}) is non-negative, the ascent direction can be simplified as  
$d =-\nabla_{\theta} F_R(r;\theta)\big|_{r=q(\alpha;\theta)}$ in searching for the optimum.
The likelihood ratio technique in (\ref{pf.eq2}) can be applied analogously to derive the gradient $\nabla_{\theta} F_R(r;\theta)$ as below:
\begin{align*}
\nabla_{\theta} F_R&(r;\theta)=\nabla_{\theta} \mathbb{E}[\mathbf{1}\{R\leq r\}]=\nabla_{\theta} \mathbb{E}[\mathbf{1}\{U(\tau)\leq r\}]\\
&=\nabla_{\theta}\int_{\Omega_{\tau}}\mathbf{1}\{U(\tau)\leq r\}\Pi(\tau;\theta)d\tau\\
&=\mathbb{E}[\mathbf{1}\{U(\tau)\leq r\}\nabla_{\theta}\log\Pi(\tau;\theta)]\\
&=\mathbb{E}\bigg[\mathbf{1}\{U(\tau)\leq r\}\sum_{t=0}^{T-1}\nabla_{\theta}\log\pi(a_t|s_t;\theta)\bigg]\\
&\approx\frac{1}{N}\sum_{n=0}^{N-1}\mathbf{1}\{U(\tau^n)\leq r\}\sum_{t=0}^{T-1}\nabla_{\theta}\log\pi(a_t^n|s_t^n;\theta).
\end{align*}
Denote $$D(\tau;\theta,r)=-\mathbf{1}\{U(\tau)\leq r\} \sum_{t=0}^{T-1}\nabla_{\theta}\log\pi(a_t|s_{t};\theta).$$ We propose a two-time-scale iterative algorithm as follows:
\begin{align}
    q_{k+1}&=q_k + \beta_k(\alpha-\mathbf{1}\{U(\tau^{k})\leq q_{k})\}),\label{iter_q}\\
    \theta_{k+1}&=\varphi(\theta_k+\gamma_k D(\tau^k;\theta_k,q_k)\label{iter_theta}),
\end{align}
where $\varphi(\cdot)$ is a projection function to guarantee $\theta_k\in\Theta$. Recursion (\ref{iter_q}) is used to track the quantile of the current policy $\pi(\cdot|\cdot;\theta_k)$, i.e., a one-step search for the root of $F_R(q;\theta_k)=\alpha$.

In practice, we divide $D(\tau^k;\theta_k,q_k)$ by the estimator of $f_R(q_k;\theta_k)$ to speed up searching for the optimum. Here we offer two feasible approaches. One is to estimate the density by the empirical reward distribution. Using the well-known Kernel Density Estimation (KDE) approach \cite{scott2015multivariate,silverman2018density}, we may construct the reward density function by a batch of episodes between two steps updating the parameters. 
An alternative approach is to approximate the indicator $\mathbf{1}\{\cdot\}$ by a smooth function that can be directly differentiated. For example, we have
\begin{align*}
    f_R(r;\theta)&=\frac{\partial\mathbb{E}[\mathbf{1}\{R\leq r\}]}{\partial r}\approx\frac{\partial\mathbb{E}\left[\sigma(r;R)\right]}{\partial r} =\mathbb{E}\left[\sigma'(r;R)\right],
\end{align*}
where $\sigma(\cdot)$ is the sigmoid function defined as 
\begin{align*}
\sigma(r;R)=\frac{1}{1+e^{-(r-R)}}.
\end{align*}
Then $\frac{1}{N}\sum_{n=0}^{N-1}\sigma'(q_k;U(\tau_k^{n}))$ can be used as an estimator of $f_R(q_k;\theta_k)$. We present the pseudo code of QPO in Algorithm \ref{alg.qpo} as follows.

\begin{algorithm}[h]
    \caption{Quantile-Based Policy Optimization (QPO)}
    \label{alg.qpo}
\begin{algorithmic}[1]
    \STATE {\bfseries Input:} Policy network $\pi(\cdot|\cdot;\theta)$, quantile parameter $\alpha\in(0,1)$, and batch size $N$.
    \STATE {\bfseries Initialize:} Policy parameter $\theta_0\in\Theta$ and quantile estimator $q_0\in\mathbb{R}$.
    \FOR{$k=0,\cdots,K-1$}
    \STATE Generate $N$ episodes $\{\tau^k_n\}_{n=0}^{N-1}$ following policy $\pi(\cdot|\cdot;\theta_k)$;
    \STATE $q_{k+1}\leftarrow q_k + \beta_k\big(\alpha-\frac{1}{N}\sum_{n=0}^{N-1}\mathbf{1}\{U(\tau^k_n)\leq q_{k}\}\big)$;
    \STATE $\theta_{k+1}\leftarrow \varphi\bigg(\theta_k+\gamma_k \frac{\sum_{n=0}^{N-1}D(\tau^k_n;\theta_k,q_k)}{\sum_{n=0}^{N-1}\sigma'(q_k;U(\tau^k_n))}\bigg)$.
    \ENDFOR
    \STATE {\bfseries Output:} Trained policy network $\pi(\cdot|\cdot;\theta_{K})$.
\end{algorithmic}
\end{algorithm}

\subsection{Quantile Optimization of PPO Style}
To improve data utilization efficiency and robustness, we propose a variant of our QPO algorithm by using an importance sampling technique inspired by PPO.
We consider a pair of policy networks $\pi(\cdot|\cdot;\theta)$ and $\pi(\cdot|\cdot;\tilde{\theta})$, where only $\pi(\cdot|\cdot;\tilde{\theta})$ is used to interact with the environment. In each epoch, a batch of episodes are generated by following policy $\pi(\cdot|\cdot;\tilde{\theta})$. Then, the target policy $\pi(\cdot|\cdot;\theta)$ is updated according to the episodes generated by $\pi(\cdot|\cdot;\tilde{\theta})$, and the parameter $\tilde{\theta}$ is updated by the latest value of $\theta$.

Denote the importance sampling ratio as
\begin{align}
     &\rho(\tilde{a}_t,\tilde{s}_t)=\frac{\pi(\tilde{a}_t|\tilde{s}_t;\theta)}{\pi(\tilde{a}_t|\tilde{s}_t;\tilde{\theta})},\nonumber\\ &\rho(\tilde{\tau})=\frac{\Pi(\tilde{\tau};\theta)}{\Pi(\tilde{\tau};\tilde{\theta})}=\prod_{t=0}^{T-1}\rho(\tilde{a}_t,\tilde{s}_t)\label{rho},
\end{align}
where the trajectory $\tilde{\tau}=\{\tilde{s}_0,\tilde{a}_0,\cdots,\tilde{s}_{T-1},\tilde{a}_{T-1},\tilde{s}_{T}\}$ is generated by $\pi(\cdot|\cdot;\tilde{\theta})$.
By noticing $\nabla_x f(x)=f(x)\nabla_x\log f(x)$ and $F_R(r;\theta)=\mathbb{E}_{\tilde{\tau}\sim\Pi(\cdot;\tilde{\theta})}[\mathbf{1}\{U(\tilde{\tau})\leq r\}\rho(\tilde{\tau})]$, we have
$$\nabla_{\theta} F_R(r;\theta)=\mathbb{E}_{\tilde{\tau}\sim\Pi(\cdot;\tilde{\theta})}[\rho(\tilde{\tau})\mathbf{1}\{U(\tilde{\tau})\leq r\}\nabla_{\theta}\log\Pi(\tilde{\tau};\theta)].$$
Thus, we can rewrite recursions (\ref{iter_q}) and (\ref{iter_theta}) as follows:
\begin{align}
    q_{k+1}&=q_k + \beta_k(\alpha-\rho(\tilde{\tau}^{k})\mathbf{1}\{U(\tilde{\tau}^{k})\leq q_{k})\}),\label{iter_q'}\\
    \theta_{k+1}&=\varphi(\theta_k+\gamma_k \rho(\tilde{\tau}^{k})D(\tilde{\tau}^{k};\theta_k,q_k)\label{iter_theta'}).
\end{align}
In practice, the gradient term $\rho(\tilde{\tau}^{k})D(\tilde{\tau}^{k};\theta_k,q_k)$ can be calculated stepwisely, which is computationally more efficient. If  $-\mathbf{1}\{U(\tilde{\tau}^{k})\leq q_{k}\}$ is used as the advantage function $A_t^k$ for each action $\tilde{a}_t^k$ in the trajectory $\tilde{\tau}^k$, then the searching direction in recursion (\ref{iter_theta'}) can be replaced by  $$\hat{d}_k=\sum_{t=0}^{T-1}\rho(\tilde{a}_t^k,\tilde{s}_t^k)A_t^k\nabla_{\theta}\pi(\tilde{a}_t^k|\tilde{s}_t^k;\theta_k),$$ which can be obtained by differentiating the following surrogate objective:
\begin{align*}
    \hat{\mathbb{E}}\bigg[\sum_{t=0}^{T-1}\rho(\tilde{a}_t,\tilde{s}_t)A_t\bigg]=\hat{\mathbb{E}}\bigg[-\sum_{t=0}^{T-1}\rho(\tilde{a}_t,\tilde{s}_t)\mathbf{1}\{U(\tilde{\tau}^{k})\leq q_{k}\}\bigg].
\end{align*}
To constrain the difference between $\theta$ and $\tilde{\theta}$, we introduce a clip function $\text{clip}(x,x^{-},x^{+})$, where $x^{-}$ and $x^{+}$ are the lower and upper truncation bounds. The clipped surrogate objective is
\begin{align}
    \hat{\mathbb{E}}_{\tilde{\tau}\sim\Pi(\cdot;\tilde{\theta})}\bigg[\sum_{t=0}^{T-1}&\min\{\rho(\tilde{a}_t,\tilde{s}_t)A_t,\nonumber\\
    &\text{clip}(\rho(\tilde{a}_t,\tilde{s}_t),1-\varepsilon,1+\varepsilon)A_t\}\bigg].\label{sur_obj}
\end{align}
The pseudo code of QPPO is presented in Algorithm \ref{alg.qppo}. To achieve a more stable performance, one may use an extra baseline network $B(\tilde{s}_t|w)$ and take $A_t=-\mathbf{1}\{U(\tilde{\tau}^{k})\leq q_{k}\}-B(\tilde{s}_t|w)$ as a new advantage function, where $B(\tilde{s}_t|w)$ is updated by minimizing the MSE of $-\mathbf{1}\{u(\tilde{\tau}^{k})\leq q_{k}\}$.

\begin{algorithm}[h]
   \caption{Quantile-Based Proximal Policy Optimization (QPPO)}
   \label{alg.qppo}
\begin{algorithmic}[1]
    \STATE {\bfseries Input:} Policy network $\pi(\cdot|\cdot;\theta)$ and $\pi(\cdot|\cdot;\tilde{\theta})$,  quantile parameter $\alpha\in(0,1)$, and batch size $N$.
    \STATE {\bfseries Initialize:} Policy parameter $\theta_0^0,\ \tilde{\theta}_0\in\Theta$, and quantile estimator $q_0^0\in\mathbb{R}$.
    \FOR{$k=0,\cdots,K-1$}
    \STATE Generate $N$ episodes $\{\tilde{\tau}^k_n\}_{n=0}^{N-1}$ following policy $\pi(\cdot|\cdot;\tilde{\theta}_k)$;
    \FOR{$j=0,\cdots,J-1$}
        \STATE Use equality (\ref{rho}) to calculate $\{\rho(\tilde{\tau}_n^{k})\}_{n=0}^{N-1}$;
        \STATE Update $q_{k}^j$ to $q_{k}^{j+1}$ by recursion (\ref{iter_q'});
        \STATE Update $\theta_{k}^j$ to $\theta_{k}^{j+1}$ by maximizing clipped surrogate objective (\ref{sur_obj});
    \ENDFOR
    \STATE $q_{k+1}^{0}\leftarrow q_k^{J-1}$, $\theta_{k+1}^{0}\leftarrow \theta_k^{J-1}$, $\tilde{\theta}_{k+1}\leftarrow\theta_{k+1}^{0}$.
    \ENDFOR
    \STATE {\bfseries Output:} Trained policy network $\pi(\cdot|\cdot;\theta_{K}^0)$.
\end{algorithmic}
\end{algorithm}

\subsection{Computation Complexity}
We have proposed two new RL algorithms based on the quantile maximization criterion, which are counterparts of two corresponding mean-based algorithms, i.e., REINFORCE and PPO.
The computational cost of calculating the stochastic gradient estimate for the policy parameter is comparable to that of the mean-based algorithms.
The main difference is that our algorithms need to estimate the quantile of the current policy.
Recursion (\ref{iter_q}) or (\ref{iter_q'}) is a one-step iteration for the one-dimensional root searching problem, which is computationally very cheap.
Compared with the corresponding mean-based algorithm, the increase in computation burden of QPO and QPPO is negligible.

\section{Convergence Analysis}
In this section, we establish the global convergence of the QPO algorithm.
Let $(\Omega,\mathcal{F},P)$ be a probability space. Define the filtration generated by our algorithm $\mathcal{F}_{k}=\{\theta_0,q_0,\cdots,\theta_k,q_k\}$ for $k=0,1,\cdots$. Here we introduce some assumptions before the analysis.

\textbf{Assumptions:}

\textbf{A1:} For any $\alpha\in(0,1)$, $q(\alpha;\theta)\in C^1(\Theta)$.

\textbf{A2:} $\nabla_{\theta}F_R(q;\theta)$ is Lipschitz continuous with respect to both $q$ and $\theta$, i.e. there exists a constant $C$ such that $\Vert\nabla_{\theta}F_R(q_1;\theta_1)-\nabla_{\theta}F_R(q_2;\theta_2)\Vert\leq C\Vert(q_1,\theta_1)-(q_2,\theta_2)\Vert$ for any $(q_i,\theta_i)\in\mathbb{R}\times\Theta$, $i=1,2$.

\textbf{A3:} The step-size sequences $\{\gamma_k\}$ and $\{\beta_k\}$ satisfy
  
(a) $\gamma_k>0$, $\sum_{k=0}^{\infty}\gamma_k=\infty$, $\sum_{k=0}^{\infty}\gamma_k^2<\infty$;

(b) $\beta_k>0$, $\sum_{k=0}^{\infty}\beta_k=\infty$, $\sum_{k=0}^{\infty}\beta_k^2<\infty$;

(c) $\gamma_k=o(\beta_k)$.

\textbf{A4:} The log gradient of the neural network output with respect to $\theta$ is bounded, i.e.,
$\nabla_{\theta}\log\pi(a|s;\theta)<\infty$ for any state-action pair $(s,a)$ and parameter $\theta$.

Assumption A1 requires that the objective function is smooth enough, which is commonly assumed in continuous optimization. Assumptions A2 and A3 are standard in stochastic approximation analysis. Assumption A4 is necessary to avoid computational overflow. 

Since $\gamma_k = o(\beta_k)$ in A3(c), recursion (\ref{iter_q}) updates on a faster scale than  recursion (\ref{iter_theta}). Let 
$$g_1(q,\theta)=\alpha-F_R(q;\theta),\quad g_2(q,\theta)=-\nabla_{\theta'}F_R(q;\theta')\big|_{\theta'=\theta},$$ and we expect them to track a coupled ODE:
\begin{align}
    \dot{q}(t)=g_1(q(t),\theta(t)),\quad \dot{\theta}(t)=\tilde{\varphi}(g_2(q(t),\theta(t))), \label{ca.eq3}
\end{align}
where $\tilde{\varphi}(\cdot)$ is a projection function satisfying
\begin{align*}
  \tilde{\varphi}(g_2(q(t),\theta(t)))=g_2(q(t),\theta(t))+p(t),
\end{align*}
where $p(t)\in-C(\theta(t))$ is the vector with the smallest norm needed to keep $\theta(t)$ in $\Theta$, and $C(\theta)$ is the normal cone to $\Theta$ at $\theta$. When $\theta(t)\in\partial\Theta$, $\tilde{\varphi}(\cdot)$ projects the gradient onto $\partial\Theta$.

Intuitively, $\theta(t)$ can be viewed as static for analyzing the dynamic of the process $q(t)$. Suppose that for some constant $\bar{\theta}\in\Theta$, the unique global asymptotically stable equilibrium of the ODE 
\begin{align}
    \dot{q}(t)=g_1(q(t),\bar{\theta}) \label{ca.eq4}
\end{align}
is $q(\alpha;\bar{\theta})$. 
Recursion (\ref{iter_theta}) can be viewed as tracking the ODE
\begin{align}
    \dot{\theta}(t)=\tilde{\varphi}(g_2(q(\alpha;\theta(t)),\theta(t))).\label{ca.eq5}
\end{align}
If $\theta^{*}=\arg\max_{\theta\in\Theta}q(\alpha;\theta)$ is the unique global asymptotically stable equilibrium of this ODE, then the global convergence of QPO to $\theta^{*}$ can be proved. Therefore, we first establish the unique global asymptotically stable equilibriums for ODE (\ref{ca.eq4}) and (\ref{ca.eq5}).

\begin{lemma}\label{lem.ode1}
$q(\alpha;\bar{\theta})$ is the unique global asymptotically stable equilibrium of ODE (\ref{ca.eq4})  for all $\bar{\theta}\in\Theta$.
\end{lemma}
\proof
By definition $q(\alpha;\bar{\theta})=F_R^{-1}(\alpha;\bar{\theta})$, $q(\alpha;\bar{\theta})$ is the unique solution of $g_1(q,\bar{\theta})=0$, i.e., the unique equilibrium point of ODE (\ref{ca.eq4}).
Consider the Lyapunov function $V(x)=(x-q(\alpha;\bar{\theta}))^2$, and the derivative $\dot{V}(x)=2(x-q(\alpha;\bar{\theta}))(\alpha-F_R(x;\bar{\theta}))$ is negative for any $x\neq q(\alpha;\bar{\theta})$. Therefore, $q(\alpha;\bar{\theta})$ is global asymptotically stable by the Lyapunov Stability Theory \cite{liapounoff2016probleme}. 
\endproof

\begin{lemma}\label{lem.ode2}
If $q(\alpha;\theta)$ is strictly convex on $\Theta$, then $\theta^{*}$ is the unique global asymptotically stable equilibrium of ODE (\ref{ca.eq5}).
\end{lemma}
\proof
If $\theta^{*}\in\Theta^{\circ}$, then $\nabla_{\theta} q(\alpha;\theta^{*})=0$ and $C(\theta^{*})={0}$;  if $\theta^{*}\in\partial\Theta$, then $\nabla_{\theta} q(\alpha;\theta^{*})$ must lie in $C(\theta^{*})$, so $p(t)=-g_2(q(\alpha;\theta^{*}),\theta^{*})$. By convexity of $q(\alpha;\theta)$ on $\Theta$, $\theta^{*}$ is the unique equilibrium point. Take $V'(x)=\Vert x-\theta^{*}\Vert^2$ as the Lyapunov function, and the derivative is $\dot{V}'(x)=2(x-\theta^{*})'(g_2(q(\alpha;x),x)+p(t))$. Since $q(\alpha;\theta)$ is strictly convex, $(\theta^{*}-x)'\nabla q(\alpha;x)>q(\alpha;\theta^{*})-q(\alpha;x)>0$ for any $x\neq\theta^{*}$, which implies $(x-\theta^{*})'g_2(q(\alpha;x))<0$. Since $p(t)\in-C(x)$, we have $(x-\theta^{*})'p(t)\leq0$. Thus, $\theta^{*}$ is global asymptotically stable.
\endproof

To prove that recursions (\ref{iter_q}) and (\ref{iter_theta}) track coupled ODE (\ref{ca.eq3}), we apply the convergence theorem of the two-scale stochastic approximation as below:
\begin{theorem}\label{th.bokar}
\cite{borkar1997stochastic} Consider two coupled recursions:
\begin{align*}
    q_{k+1}&=q_k+\beta_k (g_1(q_k,\theta_k)+\varepsilon_{1,k}),\\ 
    \theta_{k+1}&=\varphi(\theta_k+\gamma_k (g_2(q_k,\theta_k)+\varepsilon_{2,k})),
\end{align*}
where $\varphi$ is a projection function, $g_1$ and $g_2$ are Lipschitz continuous, $\{\beta_k\}$ and $\{\gamma_k\}$ satisfy A3, $\{\varepsilon_{1,k}\}$ and $\{\varepsilon_{2,k}\}$ are random variable sequences satisfying
\begin{align*}
    \sum_k\beta_k\varepsilon_{1,k}<\infty,\ \sum_k\gamma_k\varepsilon_{2,k}<\infty,\quad\ a.s.
\end{align*}
If ODE (\ref{ca.eq3}) has a  unique global asymptotically stable equilibrium $\lambda(\bar{\theta})$ for each $\bar{\theta}\in\Theta$, then the coupled recursions converge to the unique global asymptotically stable equilibrium of the ODE $\dot{\theta}(t)=\tilde{\varphi}(g_2(\lambda(\theta(t)),\theta(t)))$ a.s. conditional on that the sequence $\{q_k\}$ is bounded.
\end{theorem}

Next we proves that the sequence $\{q_k\}$ is almost surely bounded under our assumptions. The conditions in Theorem \ref{th.bokar} will be verified in Theorem \ref{th.main}.

\begin{lemma}\label{lem.q}
If A1 and A3(b) hold, then the sequence $\{q_k\}$ generated by recursion (\ref{iter_q}) is bounded w.p.1, i.e., $\sup_k|q_k|<\infty$ w.p.1.
\end{lemma}

\begin{figure}[t]
	\centering
	\includegraphics[scale=0.7]{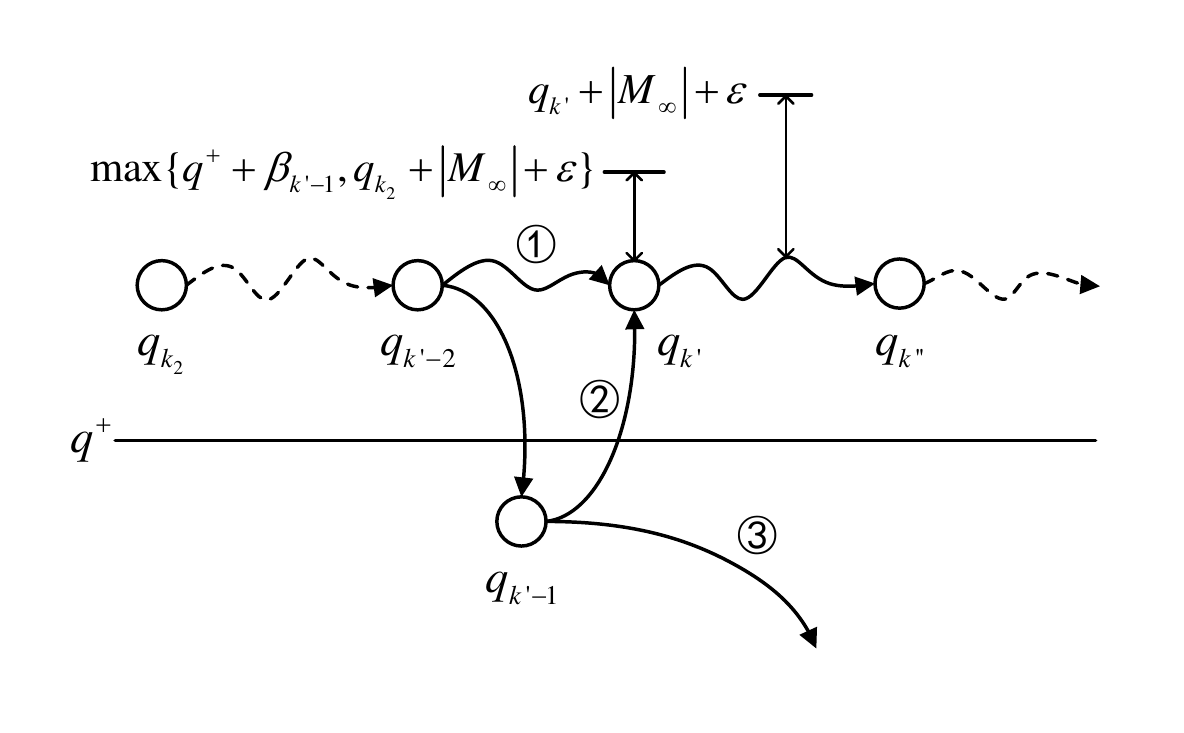}
	\caption{Illustration for possible situations of the path of $\{q_k\}$}
	\label{fig:quantile_recursion}
\end{figure}

\proof Recursion (\ref{iter_q}) can be rewritten as 
\begin{align}
    q_{k+1} = q_k +\beta_k(\alpha-F_R(q_k;\theta_k)) + \beta_k \delta_k, \label{ca.eq1}
\end{align}
where $\delta_k=F_R(q_k;\theta_k)-\mathbf{1}\{R^k\leq q_k\}$.
Let $M_k=\sum_{i=0}^k\beta_i\delta_i$. We then verify that $\{M_k\}$ is a $L^2$-bounded martingale sequence. With A3(b) and boundedness of $\mathbf{1}\{\cdot\}$, we have
\begin{align*}
    \sum_{i=0}^k\beta_i^2\delta_i^2\leq 4 \sum_{i=0}^k\beta_i^2<\infty.
\end{align*}
By noticing $\mathbb{E}[\delta_i|\mathcal{F}_i]=0$, we have
\begin{align*}
    \mathbb{E}[\ \beta_i \delta_i\beta_j \delta_j\ ]=\mathbb{E}[\ \beta_i \delta_i\mathbb{E}[\ \beta_j \delta_j|\mathcal{F}_j\ ]\ ] = 0,
\end{align*}
for all $i<j$.
Thus, $\sup_{k\geq0}\mathbb{E}[M_k^2]<\infty$. From the martingale convergence theorem \cite{durrett2019probability}, we have $M_k\rightarrow M_{\infty}$ w.p.1. Then for any $\varepsilon>0$, there exists a constant $k_0>0$ such that for any $m>n \geq k_0$, we have
\begin{align}
    \big|\sum_{i=n+1}^{m}\beta_i\delta_i\big|<\varepsilon,\quad w.p.1. \label{ca.eq2}
\end{align}
Denote $q^{+}=\sup_{\theta\in\Theta}q(\alpha;\theta)$ and $q^{-}=\inf_{\theta\in\Theta}q(\alpha;\theta)$. Since A1 holds and $\Theta$ is compact, we have $|q^{+}|<\infty$ and $|q^{-}|<\infty$. From A3(b), there exists $k_1>0$ such that for any $k \geq k_1$, $\beta_k<\varepsilon$. 
We the establish an upper bound of $\{q_k\}$. If the tail sequence $\{q_k\}_{\max\{k_0,k_1\}}^{\infty}$ keeps bounded by $q^{+}$, the boundedness of $\{q_k\}$ holds. Otherwise, let $k_2\geq\max\{k_0,k_1\}$ be the first time that $\{q_k\}_{\max\{k_0,k_1\}}^{\infty}$ rises above $q^{+}$ (i.e. $q_{k_2}> q^{+}$ and $q_{k_2-1}\leq q^{+}$), and denote a segment above $q^{+}$ of $\{q_k\}_{k_2}^{\infty}$ as $\{q_k\}_{k'}^{k''}$. 
We discuss all three possible situations as shown in Figure.\ref{fig:quantile_recursion}. On path 1, $\{q_k\}$ stays above $q^+$; on path 2, $\{q_k\}$ drops below $q^+$ first and then rises above $q^+$; and on path 3, once $\{q_k\}$ drops below $q^+$, it never rises above $q^+$.
For the situations corresponding to path 1 and path 2, the definition of $\alpha$-quantile implies $\alpha-F_R(q_k;\theta_k)<0$ for $k\in[k',k'']$. Then with equality (\ref{ca.eq1}) and inequality (\ref{ca.eq2}), we have $q_k<q_{k'}+|M_{\infty}|+\varepsilon$ for $k\in[k',k'']$. In addition, if the situation corresponds to path 1, then $q_{k'}<q_{k_2}+|M_{\infty}|+\varepsilon$ by the previous result; and if it corresponds to path 2, there holds $q_{k'}<q^{+}+\beta_{k'-1}$ by recursion (\ref{iter_q}). By the definition of $k_1$, we have a bound for both cases:
$q_{k'}<\max\{q^{+}+\beta_{k'-1},q_{k_2}+|M_{\infty}|+\varepsilon\}=q_{k_2}+|M_{\infty}|+\varepsilon$
. Thus, we have $q_k<q_{k_2}+2|M_{\infty}|+2\varepsilon$ for all $k>k_2\geq\max\{k_0,k_1\}$. Analogously, we have $q_k>q_{k_2'}-2|M_{\infty}|-2\varepsilon$ for $k>k_2'\geq\max\{k_0,k_1\}$. For the situation corresponding to path 3, the tail sequence is naturally bounded.
In summary, the conclusion has been proved.
\endproof

\begin{theorem}\label{th.main}
If A1-A4 hold and $q(\alpha;\theta)$ is strictly convex on $\Theta$, then the sequence $\{\theta_k\}$ generated by recursions (\ref{iter_q}) and (\ref{iter_theta}) converges to the unique optimal solution $\{q(\alpha;\theta^*),\theta^*\}$ of problem (\ref{pf.eq3}) w.p.1.
\end{theorem}
\proof  With $F_R(r;\theta)\in C^1(\mathbb{R})$ and A2, we have $g_1(q,\theta)$ and $g_2(q,\theta)$ are Lipschitz continuous. It has been verified in Lemma \ref{lem.q} that $M_k=\sum_{i=0}^k\beta_i\delta_i<\infty$. Denote $M'_k=\sum_{i=0}^k\gamma_i\delta'_i$, where $\delta'_i=D(\tau^i;\theta_i,q_i)+\nabla_{\theta}F_R(q_i;\theta)|_{\theta=\theta_i}$. Since $\nabla_{\theta}F_R(q;\theta)$ is Lipschitz continuous on the compact set $\Theta$ and $\{q_k\}$ is bounded as shown in Lemma \ref{lem.q}, $\nabla_{\theta}F_R(q_i;\theta)|_{\theta=\theta_i}$ is bounded. By A4, we have 
\begin{align*}
    D(\tau^i;\theta_i,q_i)&\leq\sum_{t=0}^{T-1}\nabla_{\theta}\log\pi(a_t^i|s_t^i;\theta_i)\\
    &\leq T \sup_{a,s,\theta}\nabla_{\theta}\log\pi(a|s;\theta)<\infty.
\end{align*} 
By noticing $\mathbb{E}[\delta'_i|\mathcal{F}_i]=0$, A3(a), and a similar argument in Lemma 3, we can prove $\{M'_k\}$ is a $L^2$-bounded martingale sequence, which implies that $\{M'_k\}$ is bounded w.p.1.

With Assumption A3 and the conclusions in Lemmas \ref{lem.ode1} and \ref{lem.q}, all conditions in Theorem \ref{th.bokar} are satisfied.
Therefore, it is almost sure that recursions (\ref{iter_q}) and (\ref{iter_theta}) converge to the unique global asymptotically stable equilibrium of ODE (\ref{ca.eq5}),
which is the optimal solution of problem (\ref{pf.eq3}) by the conclusion of Lemma \ref{lem.ode2}.
\endproof

\section{Experiments}

In this section, we conduct simulation experiments on different RL tasks to compare our QPO and QPPO with baseline RL algorithms REINFORCE and PPO. By experiments, we find that the SPSA-based algorithms can not train neural networks in the deep RL settings, so they are not compared with our methods.

The selection of $\{\beta_k\}$ will affect the convergence speed of the algorithm.
We propose an adaptive updating approach for $\{\beta_k\}$:
\begin{align*}
    \beta_k = c_0 k^{-\lambda} \text{clip}(|q_k|,c_1,c_2),
\end{align*}
where $\lambda\in(0.5,1)$, and $c_0$, $c_1$ and $c_2$ are positive constants. The chosen $\{\beta_k\}$ satisfies the assumption A3(b). The rationale of this learning rate is to scale up the searching step by the magnitude of the current quantile estimate clipped by some upper-and-lower bounds. In practice, to speed up searching further, $k^{-\lambda}$ can be replaced by a factor that decays at a constant rate after certain number of iterations.

\subsection{Toy Example: Zero Mean}
In a simulation environment with $T$ time steps, the reward of an agent at the $t$-th step is 
\begin{align*}
    R_t=X_t+h(\delta_t)(Y_t-\frac{1}{2}),\ X_t\sim\mathcal{N}(0,1),\ Y_t\sim B(1,\frac{1}{2}),
\end{align*}
where $\delta_t$ is an implicit system parameter vector controlled by the agent, and $h(\delta_t)$ is an explicit state vector that can be observed.
Then $R_t$ follows a bimodal distribution with zero mean. Therefore, optimizing the expected cumulative reward is ineffective, whereas the quantile performance can be optimized by controlling $\delta_t$.
\begin{figure}[h]
	\centering
	\includegraphics[scale=0.45]{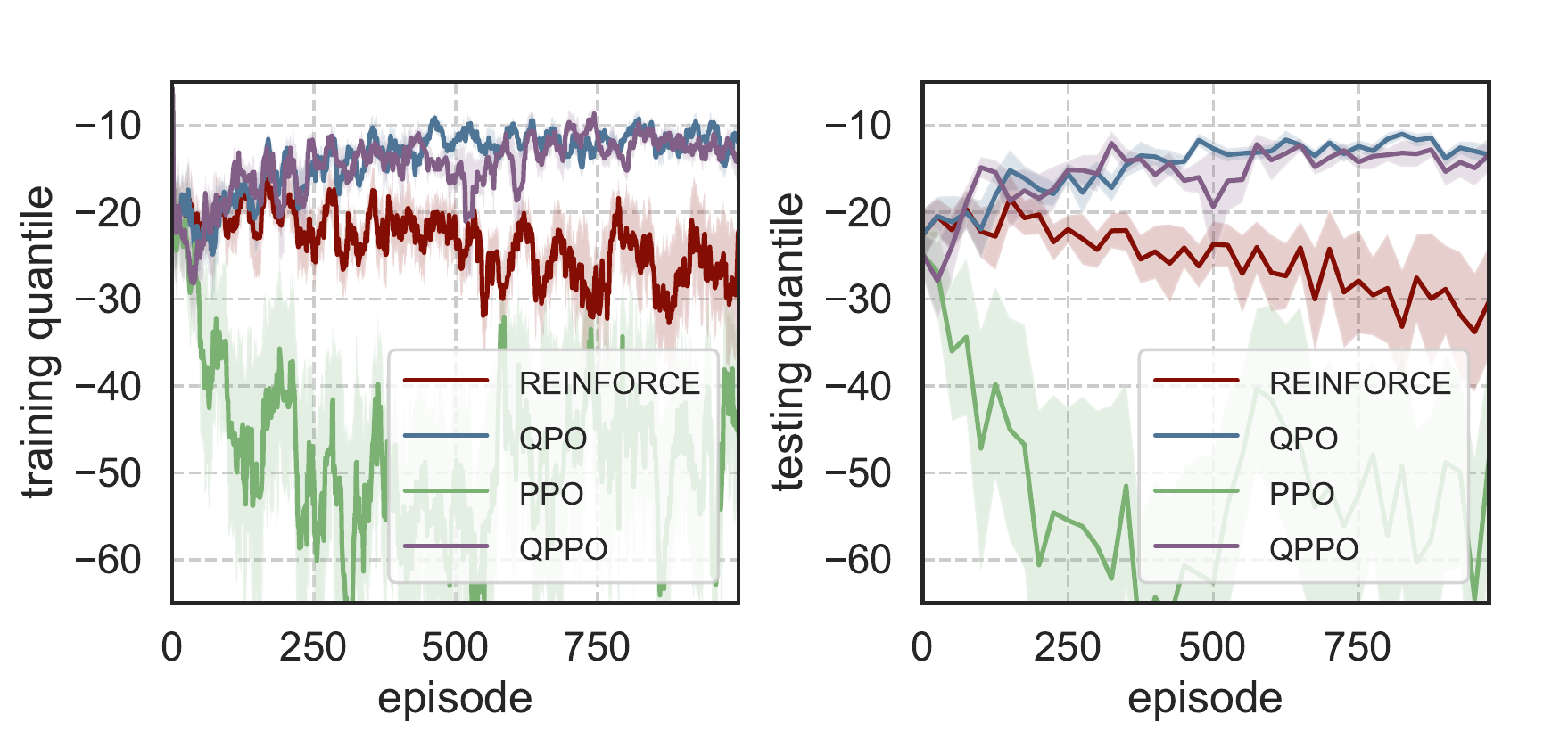}
	\caption{Learning curves for quantile ($\alpha=0.1$) of REINFORCE, QPO, PPO and QPPO in the Zero Mean example estimated by 10 independent experiments.}
	\label{fig:ZeroMean}
\end{figure}

The learning curves of REINFORCE, QPO, PPO and QPPO are presented in Figure \ref{fig:ZeroMean}.  The policy is represented by a neural network consisting of two hidden layers with each containing 32 neurons. The initial learning rate is set as $5\times 10^{-4}$, and it is multiplied by a decay factor 0.7 every 400 episodes. On the left hand side of the figure, we present the quantile estimated by total rewards in past 20 episodes. On the right hand side of the figure, we present the true quantiles estimated by 50 replications per 25 episodes. We can see that QPO and QPPO significantly improve the agent's quantile performance. In contrast, the learning curves of RINFORCE and PPO keep oscillating around a low level.

\subsection{Stock Trading with Simulated Prices: Sim Stock}
We consider a simplified stock trading problem with $N$ simulated stock prices and $T$ time steps, where high returns are associated with high risks.
The price of every stock is generated independently beforehand by a model in \cite{moody2001learning}. For the $n$-th stock, the log price $x_{t+1}$ at time $t+1$ is 
\begin{align*}
    y_{t+1}^n &= \eta_1 y_t^n + \varepsilon_t^n\\
    x_{t+1}^n &= x_{t}^n + y_{t}^n + \eta_2 v_t^n,
\end{align*}
where $\eta_1$ and $\eta_2$ are constants, $\varepsilon_t^n$ and $v_t^n$ are noises which follow a bimodal distribution, more specifically, a summation of a zero-mean normal random variable and zero-mean binomial random variable. A bimodal distribution can capture distinctive statistical characteristics of the price movements under bull and bear market scenarios. 
Then the  price sequences $\{p_t^n\}$ are simulated by 
\begin{align*}
    p_t^n=\exp\bigg(\frac{x_t^n}{\max_t x_t^n -\min_t x_t^n}\bigg).
\end{align*}
The agent initially holds a random portfolio with a fixed total value $1$ and can control the proportion of cash invested into each stock.
At every time step, the agent first makes investment decisions, and calculates the total value of the portfolio under current stock prices and that under new stock prices.
The agent is rewarded by the differences of two total values. In addition, cash held by the agent offers  $0.2\%$ risk-free return per time step. The observation state contains stock prices at the past few steps and the current portfolio.

\begin{figure}[h]
	\centering
	\includegraphics[scale=0.45]{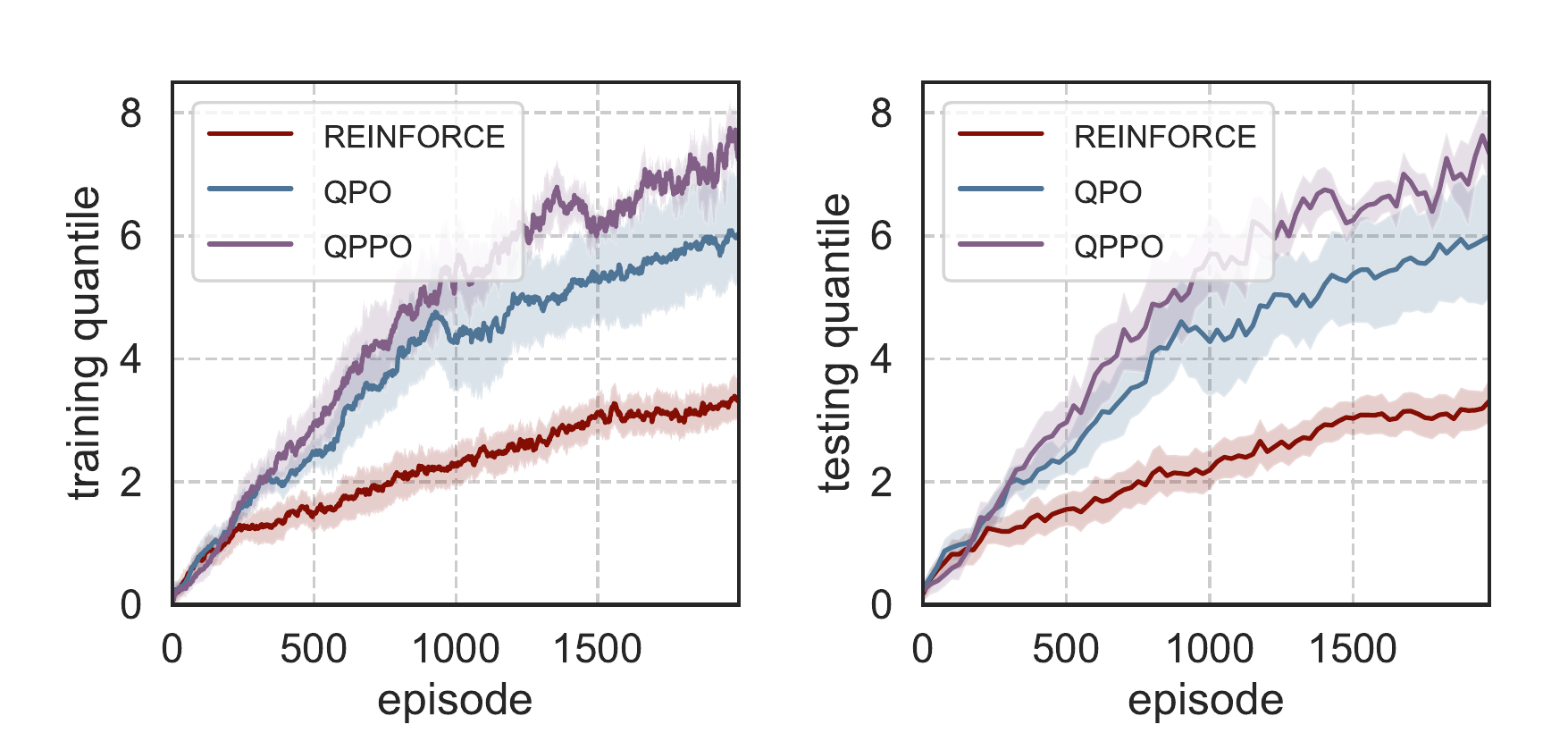}
	\caption{Learning curves for quantile ($\alpha=0.3$) of REINFORCE, QPO and QPPO in the Sim Stock example estimated by 5 independent experiments.}
	\label{fig:SimStock}
\end{figure}

The learning curves of REINFORCE, QPO and QPPO are presented in Figure \ref{fig:SimStock}.  The policy and the baseline are represented by neural networks consisting of two hidden layers with each containing 64 neurons. The initial learning rate is set as $5\times 10^{-4}$, and it is multiplied by a decay factor 0.7 every 500 episodes. The agent invests into 8 stocks in 200 time steps and can look back at stock prices within the past 5 steps. On the left hand side of the figure, we present the quantile estimated by total rewards in past 20 episodes. On the right hand side of the figure, we present the true quantiles estimated by 20 replications per 25 episodes. 

Better quantile performance of the portfolio indicates that it is more robust under extreme market scenarios.
The two quantile-based algorithms QPO and QPPO achieve much superior quantile performances than the mean-based algorithm REINFORCE in this example.  In addition, QPPO converges faster than QPO.

\subsection{Stock Trading with Real Prices: Dow Stock}
We replace the simulated stock prices with the real data of Dow Jones 30 in \cite{yang2020deep} and use the metric 'prccd' as a proxy of daily prices for the real stock market. We conduct experiments under the same setting as in simulated stock prices.

\begin{figure}[h]
	\centering
	\includegraphics[scale=0.45]{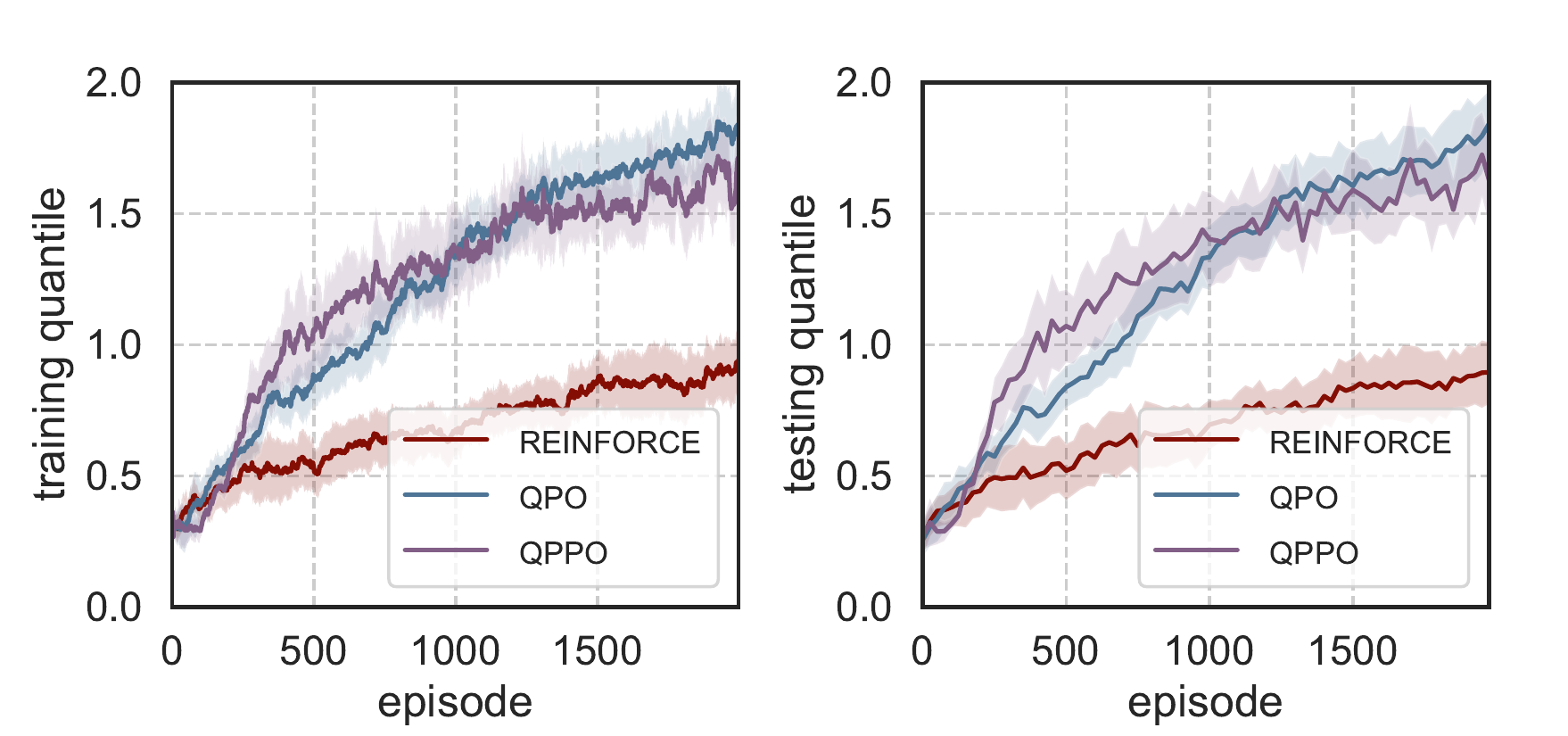}
	\caption{Learning curves for quantile ($\alpha=0.3$) of REINFORCE, QPO, PPO and QPPO in the Dow Stock example estimated by 5 independent experiments.}
	\label{fig:DowStock}
\end{figure}

The learning curves for REINFORCE, QPO and QPPO are presented in Figure \ref{fig:SimStock}. In this example, QPO and QPPO still outperform the REINFORCE algorithm. QPPO converges faster than QPO at the beginning, but it lags slightly behind the latter at the end. This phenomenon may be attributed to  the baseline network introduced in QPPO. Although baselines can speed up the convergence of policy-based algorithms, they also introduce biases. Unlike the mean-based baseline in PPO, the quantile-based baseline is not centralized so that it tends to be more difficult to learn. 

\section{Conclusion}
In this paper, we propose a QPO algorithm and its variant QPPO for RL  with a quantile criterion under a  policy optimization framework. To the best of our knowledge, this is the first deep RL algorithm that directly optimizes quantile performance.
We prove that QPO converges to the global optimum under certain conditions. The numerical experiments show that our proposed algorithms are effective for optimizing policy with a quantile criterion. 
In future work, the criterion can be extended to the distortion risk measure which is more general than quantiles. How to combine the quantile criterion with more effective policy optimization algorithms also deserves further study.

%%%%%%%%% REFERENCES
{\small
\bibliographystyle{unsrt}
\bibliography{main}
}

\end{document}